\documentclass[conference]{IEEEtran}
\IEEEoverridecommandlockouts
\usepackage{cite}
\usepackage{amsmath}
\usepackage{amssymb}
\usepackage{algorithm}
\usepackage{bbm}
\usepackage{algorithmic}
\usepackage{graphicx}
\usepackage{textcomp}
\usepackage{xcolor}
\usepackage{verbatim}
\usepackage[skip=0pt,font=footnotesize]{caption}
\def\BibTeX{{\rm B\kern-.05em{\sc i\kern-.025em b}\kern-.08em
    T\kern-.1667em\lower.7ex\hbox{E}\kern-.125emX}}

\usepackage{times}
\usepackage{epsfig}
\usepackage{bbm}
\usepackage{booktabs}
\usepackage{algorithm}
\usepackage{algorithmic}
\usepackage [english]{babel}
\usepackage [autostyle, english = american]{csquotes}
\MakeOuterQuote{"}
\DeclareMathOperator*{\argmin}{argmin}

\newlength\myindent
\newcommand\bindent{%
	\begingroup
	\setlength{\itemindent}{\myindent}
	\addtolength{\algorithmicindent}{\myindent}
}
\newcommand\eindent{\endgroup}

\newtheorem{theorem}{Theorem}[section]

\begin{document}

\title{Harnessing Adversarial Distances to Discover High-Confidence Errors}

\author{\IEEEauthorblockN{Walter Bennette}
\IEEEauthorblockA{\textit{Information Directorate} \\
\textit{Air Force Research Lab}\\
Rome, NY \\
walter.bennette.1@us.af.mil}
\and
\IEEEauthorblockN{Karsten Maurer}
\IEEEauthorblockA{\textit{Department of Statistics} \\
\textit{Miami University}\\
Oxford, OH \\
maurerkt@miamioh.edu}
\and
\IEEEauthorblockN{Sean Sisti}
\IEEEauthorblockA{\textit{Information Directorate} \\
\textit{Air Force Research Lab}\\
Rome, NY \\
sean.sisti@us.af.mil}
}

\maketitle

\begin{abstract}

Given a deep neural network image classification model that we treat as a black box, and an unlabeled evaluation dataset, we develop an efficient strategy by which the classifier can be evaluated.  Randomly sampling and labeling instances from an unlabeled evaluation dataset allows traditional performance measures like accuracy, precision, and recall to be estimated.  However, random sampling may miss rare errors for which the model is highly confident in its prediction, but wrong.  These high-confidence errors can represent costly mistakes, and therefore should be explicitly searched for.  Past works have developed search techniques to find classification errors above a specified confidence threshold, but ignore the fact that errors should be expected at confidence levels anywhere below 100\%.  In this work, we investigate the problem of finding errors at rates greater than expected given model confidence.  Additionally, we propose a query-efficient and novel search technique that is guided by adversarial perturbations to find these mistakes in black box models.  Through rigorous empirical experimentation, we demonstrate that our Adversarial Distance search discovers high-confidence errors at a rate greater than expected given model confidence.
\end{abstract}


\section{Introduction}

Given a deep neural network image classification model that we treat as a black box, and an unlabeled evaluation dataset, it is necessary to have an efficient strategy to evaluate the classifier.  For example, if a physician is teamed with some black-box diagnostic tool, it would be prudent for the physician to evaluate the tool before utilizing it in practice.  A desirable evaluation procedure should be respectful of the physician's time and effort, but help reveal the strengths and weaknesses of the model.  

One strategy to evaluate a black box model with an unlabeled evaluation dataset is to randomly sample and label instances from the dataset, and estimate traditional performance measures like accuracy, precision, and recall.  Another strategy is to sample low confidence predictions to discover areas where the model is prone to error. However, these strategies may miss errors for which the model is highly confident in its prediction, but wrong.  These high-confidence errors can represent costly mistakes (e.g misdiagnosis), and therefore should be explicitly searched for.  

In this paper, we propose a novel and query-efficient approach for guiding a human, or oracle, to high-confidence classification errors made by black box image classification models.  Specifically, we propose a search that leverages small perturbations to an image to help identify instances within an unlabeled evaluation dataset for which the classification model has high confidence in its prediction, but is wrong.  These perturbations are similar to those of recent developments in adversarial images.  Special attention is devoted to ensure that the developed technique is applicable to black box classifiers where specifics of the model's training data and architecture may be unknown.

High-confidence errors can be interpreted as blind spots to a classification model \cite{Attenberg2015}.  These high-confidence errors can be caused by dataset shift during use \cite{Sugiyama2017}, dataset bias during training \cite{Stock2017}, overfitting, and other reasons for poor model performance.  For example, \cite{Lakkaraju2016} describes a classification model learned from a biased dataset of dogs with dark fur and cats with light fur.  When used for inference, this model is highly confident that dogs with light fur belong to the cat class. Discovering that dogs with light fur can be misclassified with high confidence reveals a weakness of the classifier.

Previous efforts have designed search techniques to help discover high-confidence errors in an unlabeled evaluation set by searching for errors above a confidence threshold, $\tau$ (typically set to $0.65$ for binary classification) \cite{Bansal2018, Lakkaraju2016}.  Unfortunately, these techniques ignore the logical expectation that some amount of error is expected to occur at a confidence level less than 100\%.  Meaning, 30\% of the predictions made with 70\% confidence should be errors, 20\% of the predictions made with 80\% confidence should be errors, and so on.  Therefore, existing methods may simply discover errors by chance, not by some sophisticated search procedure that leverages commonalities between errors to increase the rate of error discovery.  Instead, in this work we consider the problem of finding errors at rates greater than expected, to encourage search methods that discover something about a model's weaknesses to increase the rate of error discovery.

Contributions of this work are summarized as follows:

\begin{enumerate}
	
	\item We define the problem of finding errors within an unlabeled evaluation dataset at rates greater than what model confidence would suggest.
	
	\item We design a novel error search that utilizes adversarial perturbations to improve the chance of discovering prediction errors.
	
	\item We empirically demonstrate that our novel search procedure finds errors at a rate greater than the rate suggested by model confidence.

\end{enumerate}

The remainder of this paper is organized as follows: In Section \ref{sec:RelatedWork} we discuss existing methods used to search for high-confidence errors, and provide background on adversarial images.  In Section \ref{sec:ProblemFormulation} we formulate the problem of discovering errors at a rate that exceeds expectation given model confidence.  Then, in Section \ref{sec:Methodology}, we introduce a novel method to search for errors that leverages adversarial perturbations to glean extra information about model confidence.  Next, in Section \ref{sec:Results} and \ref{sec:Discussion}, we present experimental results and provide a discussion.  Finally, in Section \ref{sec:Conclusions}, we conclude and provide thoughts for future research.

\section{Related Work}\label{sec:RelatedWork}

In this section we discuss existing methods to search for high-confidence errors from black box classification models within an unlabeled evaluation dataset.  We review adversarial images and their relation to our proposed search technique.  We also briefly discuss model calibration.  

\subsection{High-Confidence Errors} 

Attenberg (2015) \cite{Attenberg2015} introduced the concept of searching for high-confidence errors in relation to machine learning classification models.  Here, high-confidence errors were defined to be predictions for which a classification model was highly confident, but wrong.  Works considering the search for high-confidence errors \cite{Attenberg2015, Bansal2018, Lakkaraju2016a} all follow a general structure: 1) define a utility function to describe a search's value, and 2) develop a search method to help maximize the defined utility function.

In Attenberg  (2015) \cite{Attenberg2015}, the objective was to motivate human users to find high-confidence errors.  The defined utility was a monetary value that would be paid for every high-confidence mistake that was found.  The search method was to allow the human searcher to query the model when they discovered instances they felt the model may incorrectly classify.  As a result, the human searcher developed their own search technique to try and "Beat the Machine".  Although relevant to the initial formulation of the problem, recent papers focus on algorithmic approaches to help guide an oracle to the discovery of high-confidence errors.

The first algorithmic approach to search for high-confidence errors, within an unlabeled evaluation dataset, was introduced by Lakkaraju (2017) \cite{Lakkaraju2016}.  This human-in-the-loop search defined a utility function that gave a uniform value for each discovered high-confidence error and discounted this value by the cost of the human, or oracle, to label a sampled instance (regardless if it was a high-confidence error or not).  However, the utility function is simplified for imagery as it places uniform cost for each call to the oracle.  A multi-armed bandit algorithm was then used to search through clusters in a derived feature space to find high-confidence errors.  The search is driven by tracking the average utility of a cluster, which can be viewed as the likelihood of finding a high-confidence error in that cluster. 

Bansal and Weld (2018) \cite{Bansal2018} defined a utility function to encourage the high-confidence error search to be spread throughout a derived feature space.  Given an unlabeled evaluation dataset, $X$, where $c_x$ is the confidence of a model's prediction for $x \in X$, $Q \subseteq X$ is a query set of instances to evaluate for correctness, and $Cover(x|Q)$ is a function to calculate how much an instance $x$ is covered by an error found in the query set $Q$, the utility function is then:

\begin{equation}
U = \sum_{x \in X} c_x * Cover(x|Q).
\label{eq:bw}
\end{equation}

\noindent This utility function rewards the discovery of errors that "cover" the evaluation dataset.  Here, coverage of an instance is a function of its distance to the nearest error found in the query set, where closer points yield larger values.  Note that the utility function does not directly reward the discovery of high-confidence errors, but rather rewards finding errors that are near high-confidence points.  A greedy search was then used to search through clusters of the derived feature space where the probability of each cluster containing an error was tracked.  Full details can be found in \cite{Bansal2018}.  

Maurer and Bennette (2019) \cite{Maurer2018} present an extension to \cite{Lakkaraju2016} and \cite{Bansal2018} that identifies the flaw of valuing error discovery at the rate expected given model confidence.  Meaning, the work identifies the fact that errors should be expected for confidence levels below 100\%.  The Standardized Discovery Ratio is introduced as a new measure of search performance, and compares the actual number of discovered errors to the expected number of errors given the confidence of the model's predictions.  This measure is discussed in much greater detail in Section \ref{sec:ProblemFormulation}.

\subsection{Adversarial Images}

In Convolutional Neural Networks (CNNs) an adversarial image is formed by inserting small targeted perturbations to an original image such that it is confidently misclassified by the model \cite{Goodfellow2015}.  The difference between the adversarial example and the original input is often indistinguishable to the human eye, but is still successful at fooling the classifier.  Many methods exist to create adversarial images, and they can be split into two main classes: model-based and decision-based.  

Model-based adversarial attacks leverage knowledge of the model's weights and architecture.  For example, the fast gradient sign method \cite{Goodfellow2015} relies on gradient information to create targeted perturbations to be added to the original image.  Although effective, model-based methods require model information that may not always be available with a black-box classifier.

Decision-based adversarial attacks require no knowledge of the model's weights and architecture.  Instead, they only require access to the model to predict labels for new images \cite{Brendel2017, rauber2017}.  Of particular interest is the Boundary Attack \cite{Brendel2017} which begins with a large adversarial perturbation and iteratively reduces the amount of perturbation while still remaining misclassified.  More specifically, the attack begins with an adversarial image (perhaps created through the injection of Gaussian noise) and then performs a series of steps in random directions to reduce the size of the perturbations.  Each orthogonal step is adjusted to move along the decision boundary towards the original input, with the intent to find the minimal distance between the perturbation and the original input while still being misclassified. 

Most research of adversarial images has been devoted to creating adversarial attacks or defending against adversarial attacks.  However, Stock and Cisse (2017) \cite{Stock2017} leveraged adversarial images to identify model prototypes and criticisms to help expose classifier biases.  For example, they discovered a bias in a classifier that confidently identified street lights set against a blue sky as traffic lights.  This was done by looking at model criticisms, or, images that required the least amount of perturbation to turn the image adversarial.  Additionally, Ilyas (2019) \cite{Ilyas2019} showed that image classification models discriminate instances through features that are robust and through features that are non-robust.  Robust features are highly predictive and related to the classification task as perceived by humans.  Non-robust features can also be highly predictive, but do not necessarily pertain to the human perceived classification task (in the example above, perhaps the presence of features related to the sky).  Ilyas (2019) \cite{Ilyas2019} also showed that non-robust features are more susceptible to an adversarial attack.  Meaning, a classification decision based on a non-robust feature may require less perturbation to change the prediction.  These two works hint that there may be a discrepancy between a classifier's prediction confidence and the amount of perturbation required to turn an image adversarial, and could be leveraged to help discover prediction errors.  This is explored further in Section \ref{sec:Methodology}. 

\subsection{Model Calibration}

Guo (2017) \cite{Guo2017} found that modern neural networks are poorly calibrated.  Meaning, the maximum value of the softmax layer, often taken as the confidence of the classifier's prediction, does not represent a true probability of correctness.  As stated in Guo (2017), given a model $M$, with $M(X) = (\hat{Y}, \hat P)$, where $X$ are model inputs with true labels $Y$, $\hat Y$ are model predictions, and $\hat P$ are model confidences, a perfectly calibrated model satisfies the following: 

\begin{equation}
\mathbb{P}\left(\hat{Y} = Y | \hat{P} = p \right) = p, \ \ \ \forall p \in [0,1]. 
\label{eq:calibration}
\end{equation}

\noindent Meaning, for a well calibrated model, the probability that a prediction made with $p$ confidence is correct, is equivalent to the reported confidence, $p$.  Although in practical settings perfect model calibration cannot be achieved, \cite{Guo2017} showed that temperature scaling can be used to adequately calibrate neural network models with a validation dataset.

Therefore, all of the classifiers used in our study have been calibrated using temperature scaling on a validation dataset.  The intention of this step is to ensure that the discovered overconfident errors are not an artifact of poor model calibration, but systematic errors made by the classifier for the unlabeled evaluation dataset.

\section{Problem Formulation}\label{sec:ProblemFormulation}

Lakkaraju (2017) \cite{Lakkaraju2016} defined a utility function that valued the discovery of high-confidence errors uniformly.  Bansal and Weld (2018) \cite{Bansal2018} defined a utility function that weighted the discovery of a high-confidence error by the amount of the dataset it helps "explain", or the "coverage" of the error.  This was done to discourage the search method from sampling a rich pocket of errors and ignoring the rest of the search space.  Additionally, both of these formulations defined a high-confidence error to be a classification error above a prediction confidence threshold $\tau$, where $\tau$ is set to 65\% for binary classification.  

Unfortunately, these prior formulations ignore the reasonable expectation that prediction errors should occur at confidence levels anywhere below 100\%.  Meaning, the existing search methods may simply discover errors by chance, not from some sophisticated search procedure that leverages commonalities between errors to increase the rate of error discovery.  More specifically, we argue that discovering errors at the expected rate is no more informative about the weaknesses of the model than random search, because it could be expected to find the same number of errors by randomly sampling predictions from a defined confidence range.  Instead, we should explicitly encourage the discovery of errors at rates that exceed expectations to promote search methods that uncover model weaknesses to increase their rate of discovery.  

We consider the problem of discovering high-confidence errors at rates greater than what a model's confidence would suggest, which was recently introduced by \cite{Maurer2018}.  Given a black-box classifier, $M$, with $M(x) = (\hat{y}_x, \hat{p}_x)$, where $x$ is an instance from an unlabeled evaluation set $X$, $\hat{y}_x$ is the model's prediction,  $\hat{p}_x$ is the model's confidence, and $y_x $ is the true label assigned by some oracle, the task is to find a query set of data points, $Q \subseteq X$, that maximize the Standardized Discovery Ratio (SDR).  Here SDR is defined as:

\begin{equation}
SDR = \frac{\sum_{q \in Q} \mathbbm{1}(\hat{y}_q \neq y_q)}{\sum_{q \in Q}\left(1-\hat{p}_q\right)},
\label{eq:sdr}
\end{equation}

\noindent where $|Q| = B$, $\hat{p}_q > 0.65$ for $q \in Q$, and $B$ represents the labeling budget of the oracle used to find true labels.  The SDR can then be interpreted as the number of discovered misclassifications relative to what would be expected given the confidence of the predictions.

In this formulation, a query set that leads to an SDR of one indicates that errors were discovered at the rate expected given the confidence of the predictions.  Values greater than one indicate that errors were discovered at a rate greater than expected.   While previously developed methods do not explicitly value this type of error discovery, it is still possible that errors are discovered at rates that exceed expectation.  However, this formulation allows us to explicitly value this higher rate of discovery, and maximize the value of the oracle's search.

Additionally, Theorem 3.1 shows that a query set for a model who's SDR has an expected value greater than one is an indicator of model overconfidence.  However, a query set obtained through some search procedure with an SDR  greater than one does not prove the existence of model overconfidence, because the i.i.d\ assumption of the proof is almost certainly violated.  Still, an SDR greater than one does suggest the presence of model overconfidence, and the discovered errors may provide insight to particular weaknesses.   

\begin{theorem}
	Suppose there exists a sufficiently large query set sampled i.i.d.\ from an unlabeled evaluation dataset. If the expected SDR is greater than one then there exists some level of model confidence where the probability of a correct prediction is less than the model confidence.
\end{theorem}
\begin{IEEEproof} Assume the probability of making a correct prediction at a specific model confidence is always greater than or equal to the model confidence.

	 The expected value of the SDR can be calculated as: 
	
	\[E[SDR] = \frac{E\left[\sum_{q \in Q} \mathbbm{1}(\hat{y}_q \ne y_q)\right]}{E\left[\sum_{q \in Q}\left(1-\hat{p}_q\right)\right]}.\] 
	
	\noindent The expected number of errors in the numerator can be substituted with the expectation of the true probability of error, simplifying to: 
	
	\[E[SDR] = \frac{\sum_{q \in Q} E\left[\left(1-p_q\right)\right]}{\sum_{q \in Q}E\left[\left(1-\hat{p}_q\right)\right]} \le 1 .\] 
	
	\noindent Because of our assumption we know the numerator is smaller than or equal to the denominator implying the expected SDR must be less than or equal to one.  Therefore, if the expected SDR is greater than one there must exist some model confidence that is greater than the probability of a correct prediction at that confidence level, and the model is overconfident.
\end{IEEEproof}

\section{Adversarial Distance Search}\label{sec:Methodology}

This section introduces a methodology to search for high-confidence errors that utilizes an Adversarial Distance measure to guide the search.

\subsection{Adversarial Distance}

A classifier's prediction for a specific image can be changed by strategically perturbing the pixels of that image until the classifier assigns it a different label.  These perturbations result in an adversarial image if the image has been minimally changed such that a human can not identify the difference.  In our work we call an image adversarial if it has been perturbed and changes the classifier's original prediction, even if the new prediction matches the image's true label.  Additionally, in our work, no human check is performed to verify that the image has been minimally changed. 

Given the work in \cite{Ilyas2019}, we believe deep neural network models may erroneously base some of their high-confidence predictions on non-robust features.  Using the observation of \cite{Stock2017} that some images are easier to turn adversarial than others, and the observation of \cite{Ilyas2019} that non-robust features are easily broken by adversarial attacks, we present a measure to help identify predictions that are more susceptible to adversarial attack.  Or alternatively, a measure to identify instances for which the model is potentially overly confident in its prediction due to the presence of non-robust features.  If these types of predictions are indeed more prone to error than what is suggested by the confidence of the model, selecting them for a query set would result in an SDR greater than one and reveal something about model weaknesses. 

We introduce the term Adversarial Distance to describe how much perturbation an image needs for the classifier to change its prediction in comparison to the expected amount of perturbation, as determined by predictions of similar confidence.  To begin, Mean Absolute Error (MAE) can be used to measure the mean pixel-wise difference between an adversarial image and the original image.  MAE can be calculated for two $N$x$M$ images called $a$ and $b$ as:

\begin{equation}
MAE(a,b) = \frac{1}{NM} \sum_{i=1}^N \sum_{j=1}^M |a_{(i,j)} - b_{(i,j)}|
\end{equation}

\noindent Adversarial Distance is then defined to be the difference between an image's observed MAE and expected MAE, based on confidence. 

\begin{equation}
AdvDist(x) = MAE\left(x, A(M,x) \right) - F\left(\hat{p}_x\right),
\end{equation}

\noindent where $x$ is the image for which we are calculating the Adversarial Distance, $M$ is the classifier being evaluated, $A(M,x)$ is a mechanism to alter $x$ such that $M$'s prediction is changed, and $F\left(\hat{p}_x\right)$ is a function to calculate the expected MAE based on the classifier's predictive confidence, $\hat{p}_x$, for the instance, $x$.

In our work, the mechanism $A$, used to create an adversarial image is the decision-based Boundary Attack \cite{Brendel2017} discussed in Section \ref{sec:RelatedWork}.  As a reminder, we call an image adversarial if the perturbation of the image changes the classifier's original prediction.  The intuition behind the attack, as described by Brendel (2017) \cite{Brendel2017} and repeated here, is that the algorithm begins with an image that is already adversarial (perhaps through the addition of Gaussian noise) and performs a random walk to "follow the boundary between the adversarial and the non-adversarial region such that (1) it stays in the adversarial region and (2) the distance towards the target image is reduced" \cite{Brendel2017}.  The Boundary Attack was selected to create adversarial images because it finds progressively smaller perturbations to make the image adversarial, and because it is a decision-based attack that requires no model information.  

\begin{figure}[t]
	\begin{center}
		\includegraphics[width=1\linewidth]{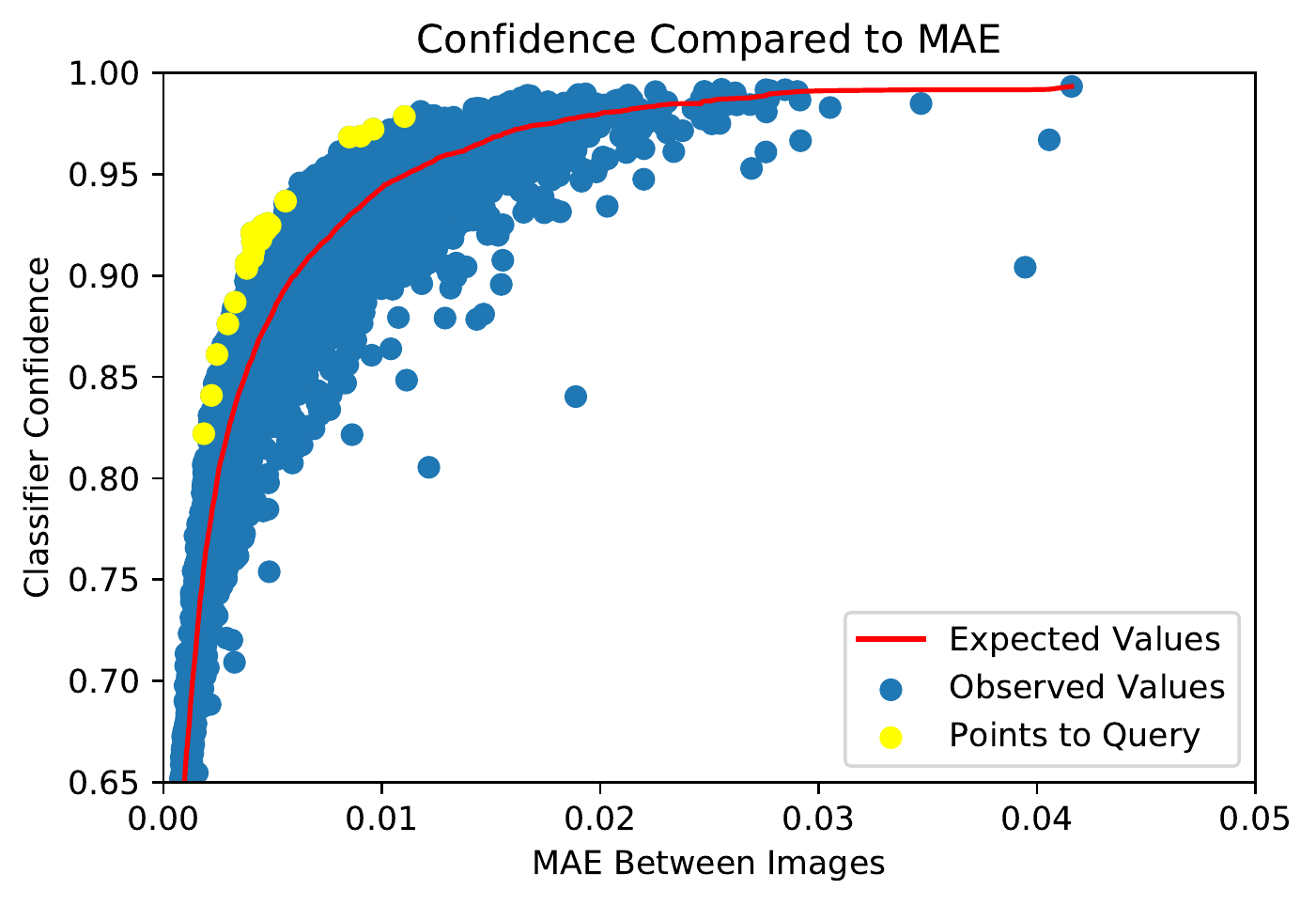}
	\end{center}
	\caption{Example LOESS curve fit between log-MAE and classifier confidence.  The horizontal distance of points from the fitted line represents the Adversarial Distance, thus the yellow points fall farthest to the left of the fitted LOESS line and have the smallest Adversarial Distances. The yellow instances would be used to query the oracle and search for errors.}
	\label{fig:curve}
\end{figure}

$F$ should provide an expected MAE given the confidence of a prediction.  Here, we calculate the MAE for every item in the evaluation set and fit a LOESS \cite{cleveland1988} regression line to estimate $F$.  MAE is the dependent variable for this LOESS line, and the classifier confidence score is the independent variable.  Figure \ref{fig:curve} shows an example LOESS for this application (built using the Kaggle13 dataset described later). Note that the horizontal distance of points from the fitted line represents the Adversarial Distance, thus the points falling farthest to the left of the fitted LOESS line will have the smallest Adversarial Distances.  Additionally, note that calculating Adversarial Distance is completely unsupervised because the true labels of the images are not needed.

\subsection{Search}

Once the Adversarial Distance has been calculated for every instance in the evaluation set, the search for high-confidence mistakes is easily defined.  Intuitively, the search queries an oracle to label the instances with the lowest Adversarial Distance.  In Figure \ref{fig:curve}, these instances are colored yellow.  Algorithm \ref{alg:search} defines the search in detail.


\begin{algorithm}
	\caption{Adversarial Distance Search}
	\label{alg:Greedy}
	\begin{algorithmic}
		\STATE {\bfseries Input:} Evaluation set $\mathbb{X}$, budget $B$, and classifier $M$
		\STATE $Q=\{\}$, instances that have been queried
		\STATE $S = \{\}$, misclassified instances
		\STATE{\bfseries For: } $b = 1, 2, ..., B$ {\bfseries do:}
		\bindent
		\STATE $q = \argmin_{x \in \mathbb{X} \ and \ x \not\in Q} AdvDist(x)$	
		\STATE $y_{q} = OracleQuery(q)$
		\STATE $Q \leftarrow Q \cup q$
		\STATE{\bfseries If:} $y_q \neq M(q)$ {\bfseries :} $S \leftarrow S \cup q$
		\eindent
		\STATE {\bfseries Return: $Q$ and $S$}
	\end{algorithmic}
	\label{alg:search}
\end{algorithm}

Algorithm 1 operates by placing the image with the lowest Adversarial Distance not already in the query set, $Q$, into the query set.  The oracle then labels the image and if the label does not match the model prediction the instance is added to the set $S$.  Once the oracle has been queried $B$ times, the search is concluded, and the set of queried points $Q$, and discovered errors, $S$, are returned for inspection.

\section{Results}\label{sec:Results}

This section introduces the experimental datasets, classifiers, evaluation procedure, and results.

\subsection{Datasets and Classifiers}

The proposed Adversarial Distance search method is evaluated using three experimental datasets.  Each dataset introduces high-confidence errors in a different way.  In line with previous works, high-confidence errors are searched for over a critical class in a binary classification problem.  Below is a description of each dataset and the mechanism by which high-confidence errors were introduced to the evaluation set.  

\textbf{Kaggle13}: The Kaggle13 dataset contains 25,000 images of cats and dogs, randomly split into equal sized train and test sets, with 1/10th of the train set reserved for validation.  The classification task is defined such that the classifier needs to determine animal type: "cat" or "dog".  High-confidence errors are introduced through dataset bias during training; black cats are removed from the training dataset. When searching for high-confidence errors, the "cat" class is the critical class.  This dataset was originally used in \cite{Lakkaraju2016} and made available by \cite{Bansal2018}.    

\textbf{CelebA}: The CelebA dataset contains 202,599 images of faces split into a predefined train, validation, and test datasets\cite{liu2015}.  We restricted our test set to 10,000 images for computational considerations.  The classification task is defined such that the classifier needs to determine gender, "male" or "female".  High-confidence errors are introduced by simulating a small dataset shift: the training set is made of RGB images and the test dataset has been converted to gray scale.  When searching for high-confidence errors, the "male" class is the critical class.  

\textbf{UT-Zap50K}: The UT-Zap50K dataset contains 50,025 images of footwear which is randomly split into a 2/3rds training set and a 1/3rd test set, with 1/10 of the train set reserved for validation \cite{yu2014, yu2017}.  The classification task is defined such that the classifier needs to identify footwear type, "not shoe" or "shoe".  Note that boots are removed from each dataset to remove easy elements of the classification task, resulting in the removal of 12,834 images.  High-confidence errors are introduced by overfitting; the classifier is trained for 75 epochs (25 epochs produces an adequate classifier).  When searching for high-confidence, the "not shoe" class is the critical class.  

A CNN with eight convolutional layers and three linear layers is used to build a classifier for each dataset.  Models are trained until the classifier stops improving on the validation dataset (with the exception of the UT-Zap50K dataset as described in the dataset description).  Furthermore, the validation datasets are used to perform temperature scaling for each classifier as recommended by \cite{Guo2017} to rectify the naturally poor calibration of CNNs.  The intent is to help ensure that any discovered high-confidence errors are not simply an artifact of poor model calibration, but a true deficiency of the model on the simulated unlabeled evaluation datasets. 

\begin{figure}[t]
	\begin{center}
		\includegraphics[width=1\linewidth]{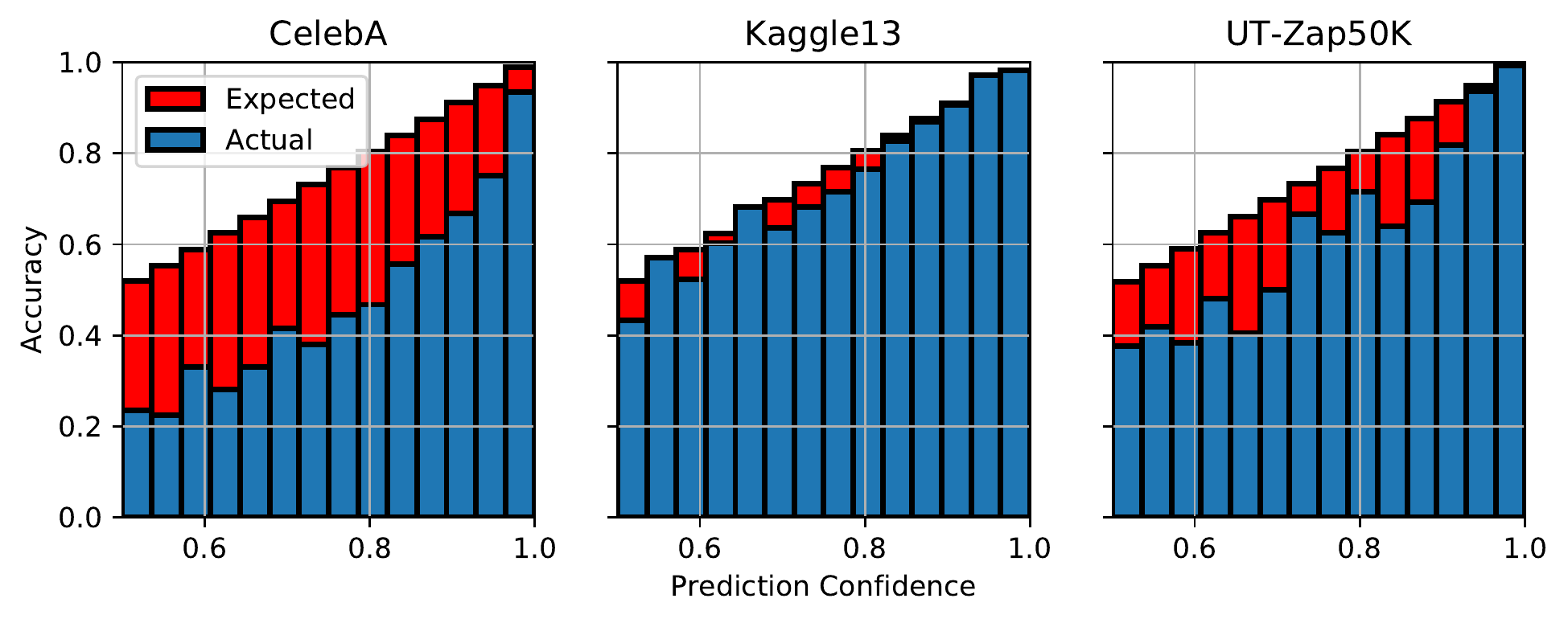}
	\end{center}
	\caption{Reliability diagram for each dataset/classifier pair. The red in each diagram indicates overconfidence. The three datasets have differing levels of overconfidence.}
	\label{fig:overconfidence}
	\label{fig:onecol}
\end{figure}

Figure \ref{fig:overconfidence} shows a reliability diagram, as described in \cite{Guo2017}, for each test dataset and temperature scaled classifier.  The reliability diagram compares expected accuracy to actual accuracy for the test dataset by using a classifier's predictions, confidences, and true labels binned for different confidence levels.  Visible red portions on the reliability diagrams represent model overconfidence, or confidence levels where more errors exist than are expected.  The reliability diagrams shown here focus exclusively on the critical class of the dataset (as identified in the dataset description), and reveal that varying levels of overconfidence are represented by these three dataset/classifier pairs.  Additionally, Table \ref{tab:Accuracy} shows the validation and test accuracy for each dataset and classifier pair.  The large drop in test accuracy for the CelebA dataset is mainly attributed to the conversion of the test dataset to gray scale; much smaller drops were observed when avoiding the dataset shift.

\begin{table}[]
	\begin{center}
		\begin{tabular}{@{}lll@{}}
			\toprule
			Dataset & Validation & Test \\ \midrule
			Celeb A & 98\% & 81\% \\
			Kaggle13 & 81\% & 81\% \\
			UT-Zap50K & 98\% & 94\% \\ \bottomrule
		\end{tabular}
	\end{center}
	\caption{Classifier performance split by dataset.}
	\label{tab:Accuracy}
\end{table}

\subsection{Evaluation}

The purpose of our evaluation is to determine if 1) a search driven by Adversarial Distance will discover diverse errors, and if 2) a search driven by Adversarial Distance can help discover a query set with an SDR greater than one, indicating that errors are discovered at a rate exceeding the rate expected given the confidence of the model's predictions.  We evaluate each component separately.

As motivated by Bansal and Weld \cite{Bansal2018}, it is desirable to discover diverse errors to avoid sampling a rich pocket of high-confidence errors \cite{Bansal2018}.  We measure the diversity, or spread, of the discovered errors as the average minimum distance from each instance in the evaluation set to an instance selected by the search. To simulate the evaluation of a black-box classifier, and to stay consistent with previous literature, spread is calculated with a feature space derived from the principal components of the evaluation set's pixel space, as we may not have access to the features used to train the classifier.  Euclidean distance is used for distance measurements.  For an evaluation set, $X$, and a query set, $Q$, spread is defined as, 

\begin{equation}
spread = \frac{\sum_{x \in \mathbf{X}} min_{q \in Q}dist(x,q)}{|\mathbf{X}|}.
\end{equation}

As previously motivated, the SDR is used to assess the quality of the query set.  SDR is defined in Equation \ref{eq:sdr} and is the ratio of discovered errors to the expected number of errors given classifier confidence.  Again, this measure provides greater insight to the quality of the search rather than defining an arbitrary threshold at which a discovered mistake is deemed valuable.

\subsection{Experiments}

The proposed Adversarial Distance search is compared to the Lakkaraju search \cite{Lakkaraju2016a}, the Bansal and Weld search \cite{Bansal2018}, a search that samples the lowest confidence predictions, and a random search.  To encourage follow on research, all of the code used to perform our experiments is available at https://github.com/afrl-ri/adversarialDistance.  Code for the Lakkaraju and Bansal and Weld search was generously made available in \cite{Bansal2018} and used for this experimentation.  Data analysis was done with R and the tidyverse packages \cite{r2017} \cite{tidy2017}.

Due to the sensitivity of the searches to the initial conditions of the unlabeled evaluation dataset, each search is run 1,000 times using a random 2,000 instance subset of the test data.  This replication simulates having 1,000 unlabeled evaluation datasets for each classifier and search method.  Each evaluation set only contains instances predicted by the classifier to belong to the critical class with confidence greater than 65\% (the threshold used in previous works to denote a high-confidence error).  Each search selects a 50 sample query set and is compared using spread and SDR.  

\begin{figure}[!htb]
	\begin{center}
		\includegraphics[width=0.9\linewidth]{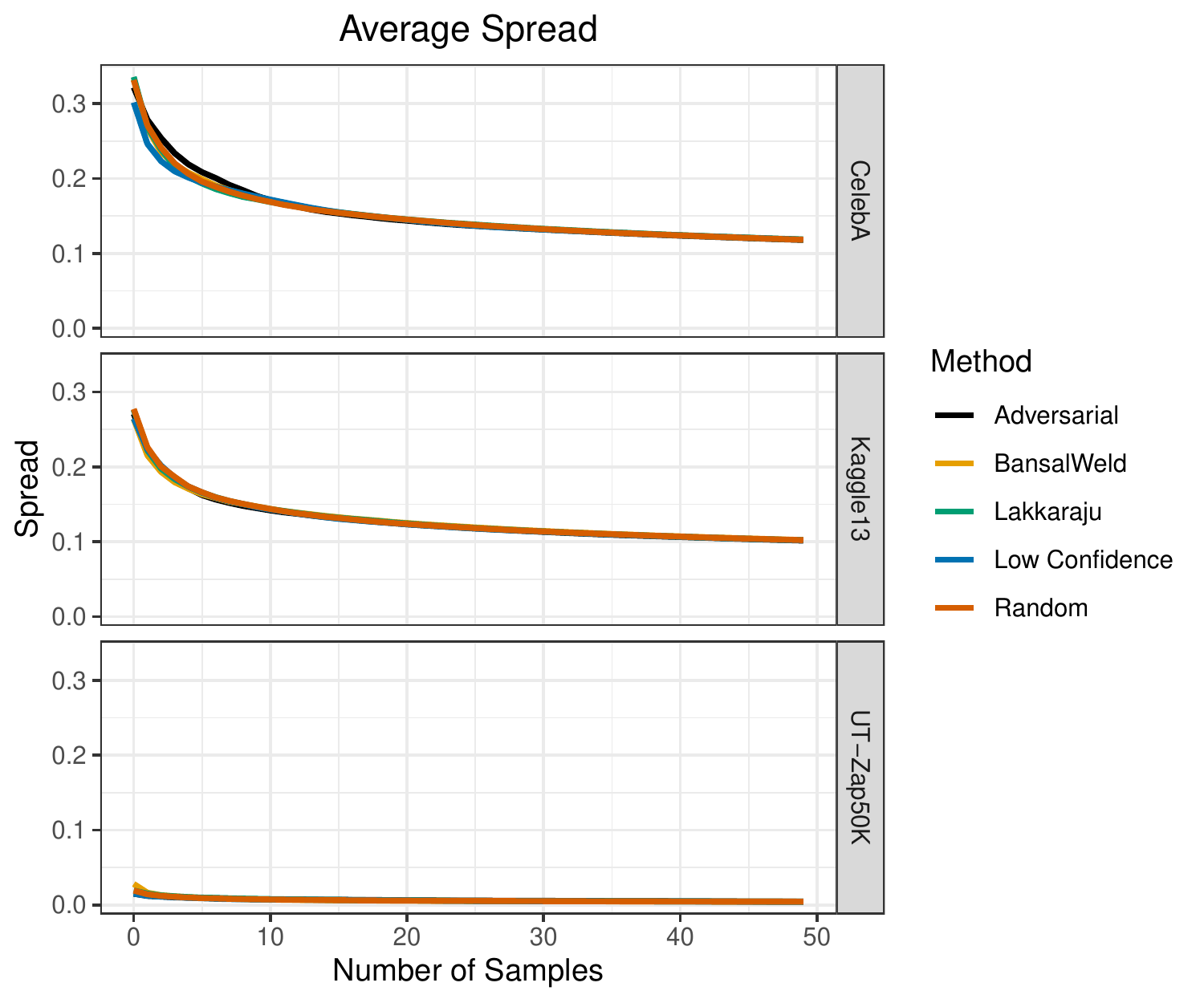}
	\end{center}
	\caption{Mean spread across 1,000 runs of the search methods.  All methods achieve a similar spread, and the spread improves as more data points are sampled.}
	\label{fig:spread}
\end{figure}

Figure \ref{fig:spread} shows the mean spread of each search over 50 queries to the oracle.  It is worth noting that all methods achieve similar spread, even in comparison to the Bansal and Weld search which is specifically designed to sample throughout the search space.  This indicates that searches are likely not getting stuck in areas with high rates of error (as previously feared), but rather are sampling throughout the search space. 

\begin{figure}[!htb]
	\begin{center}
		\includegraphics[width=0.9\linewidth]{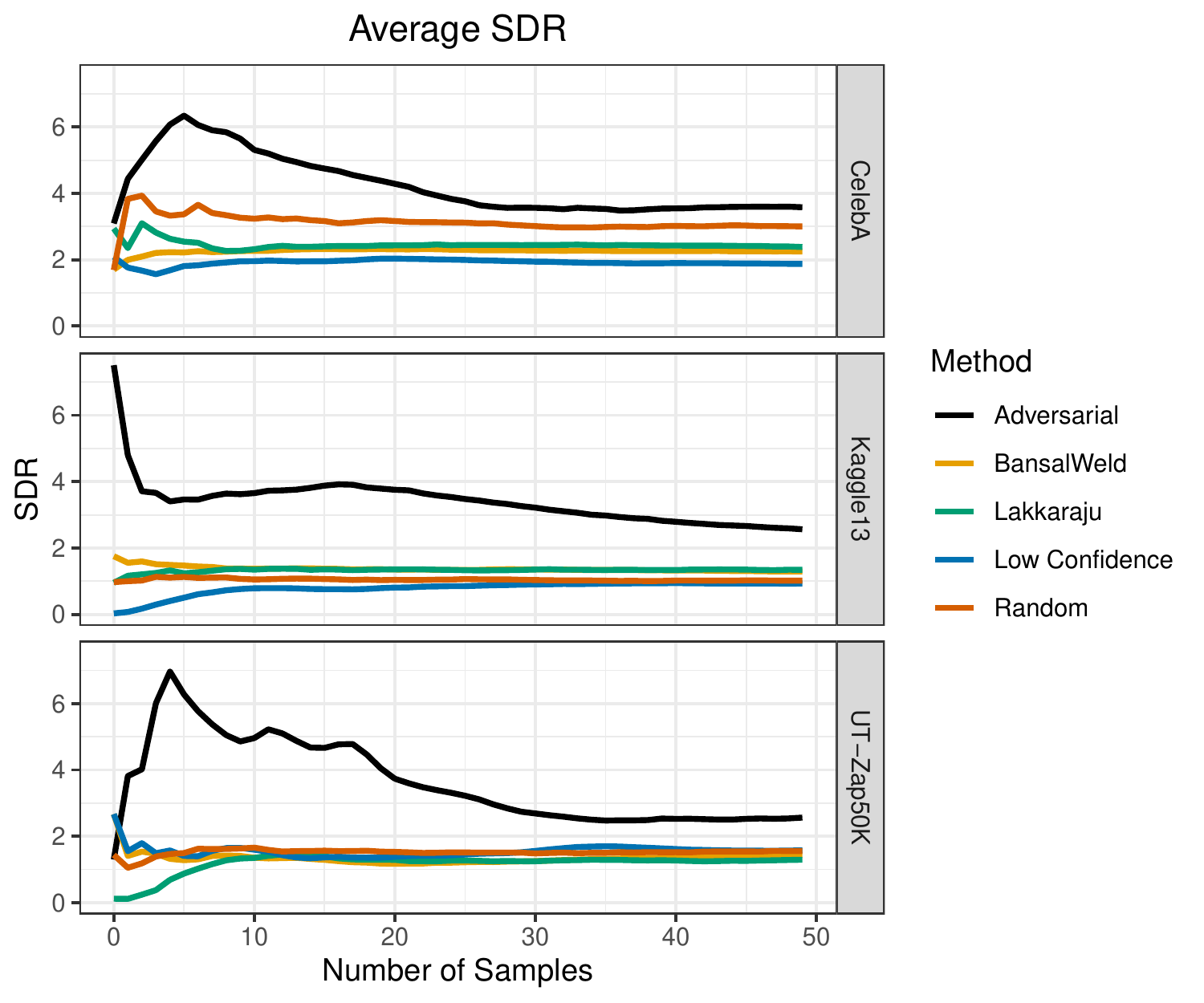}
	\end{center}
	\caption{Mean SDR across 1,000 runs of the search methods.  The Adversarial Distance method achieves the highest SDR values, meaning it has the highest rate of error discovery relative to the expected rate of error discovery.}
	\label{fig:sdr}
\end{figure}

Figure \ref{fig:sdr} shows the mean SDR of each search method over 50 queries.  The average SDR achieved by the Adversarial Distance search dominates the curves of the other methods, and indicates that this search method finds errors at rates that exceed expectations.  The other methods achieve an SDR near one for the Kaggle13 and UT-Zap50K datasets, which indicates that they are discovering errors at the rate indicated by model confidence.  For the CelebA dataset the other methods discover errors at nearly twice the rate indicated by model confidence, but this is not surprising given the amount of overconfidence shown in Figure \ref{fig:overconfidence}.  Interestingly, the performance of the Adversarial Distance search decreases as the query size increases, indicating that the density of error prone instances lessens as the adversarial distance increases.  Recommendations to alleviate this issue will be discussed in Section \ref{sec:Conclusions}.   

Of particular interest, in regards to SDR, is the performance of the Adversarial Distance search for the Kaggle13 dataset.  Recalling the reliability diagram in Figure \ref{fig:overconfidence}, there is very little overconfidence for any of these search methods to discover.  However, at 20 queries the Adversarial Distance search discovers errors at four times the rate that the model confidence would suggest, while the other methods are discovering errors at almost exactly the rate indicated by model confidence.  Even at 50 queries, the Adversarial Distance search is finding more than twice as many errors as model confidence would suggest.    

\section{Discussion}\label{sec:Discussion}


In this section, we discuss the Adversarial Distance search when considering the utility functions presented in previous works.  We then show some of the high-confidence mistakes discovered by the Adversarial Distance search and discuss what they tell us about model quality.  We also provide a discussion on why Adversarial Distance helps reveal these informative instances.

\subsection{Other Utility Functions}

\begin{figure}[!htb]
	\begin{center}
		\includegraphics[width=0.9\linewidth]{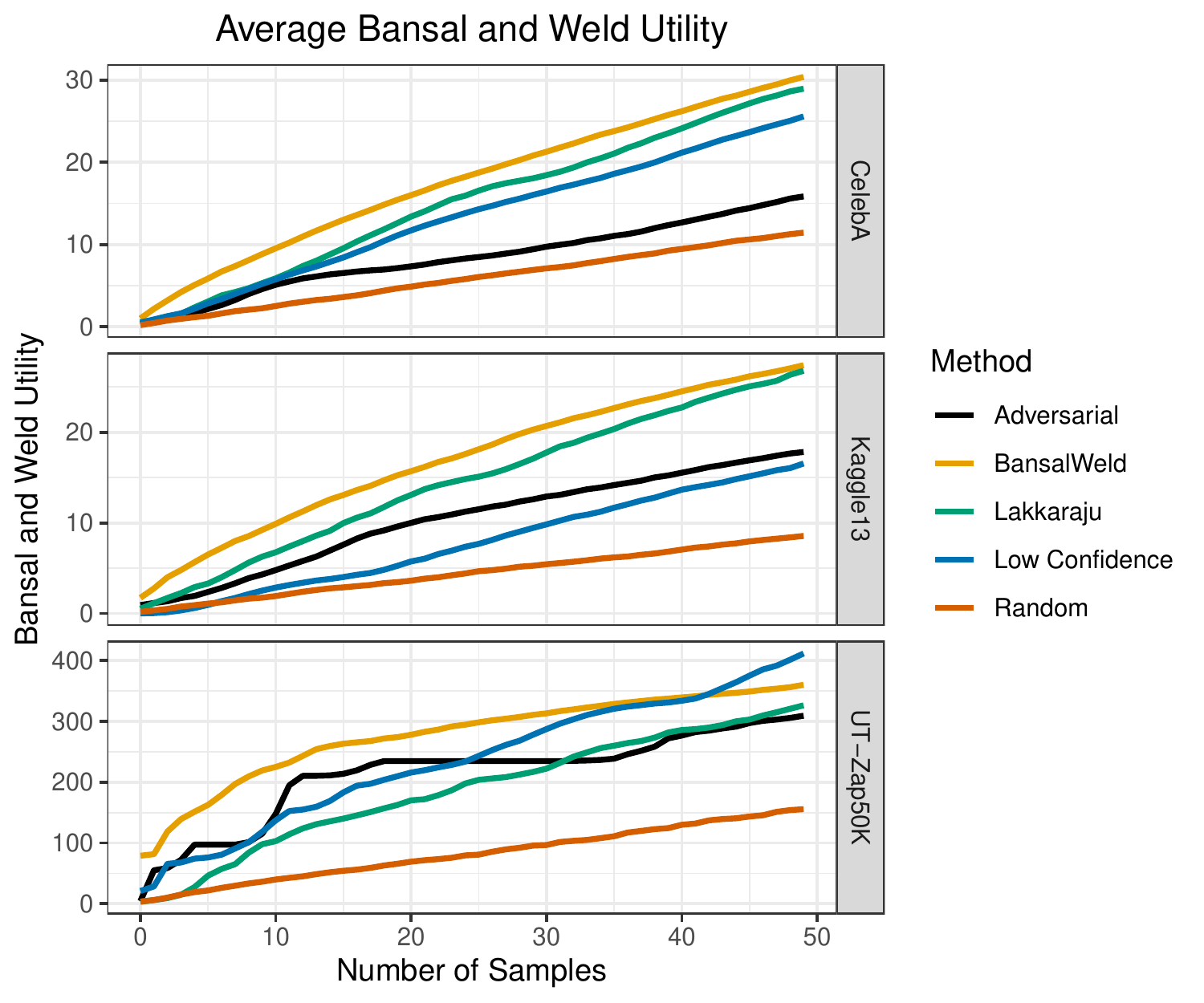}
	\end{center}
	\caption{Mean Bansal and Weld utility across 1,000 runs of the search methods.}
	\label{fig:bw}
\end{figure}

\begin{figure}[!htb]
	\begin{center}
		\includegraphics[width=0.9\linewidth]{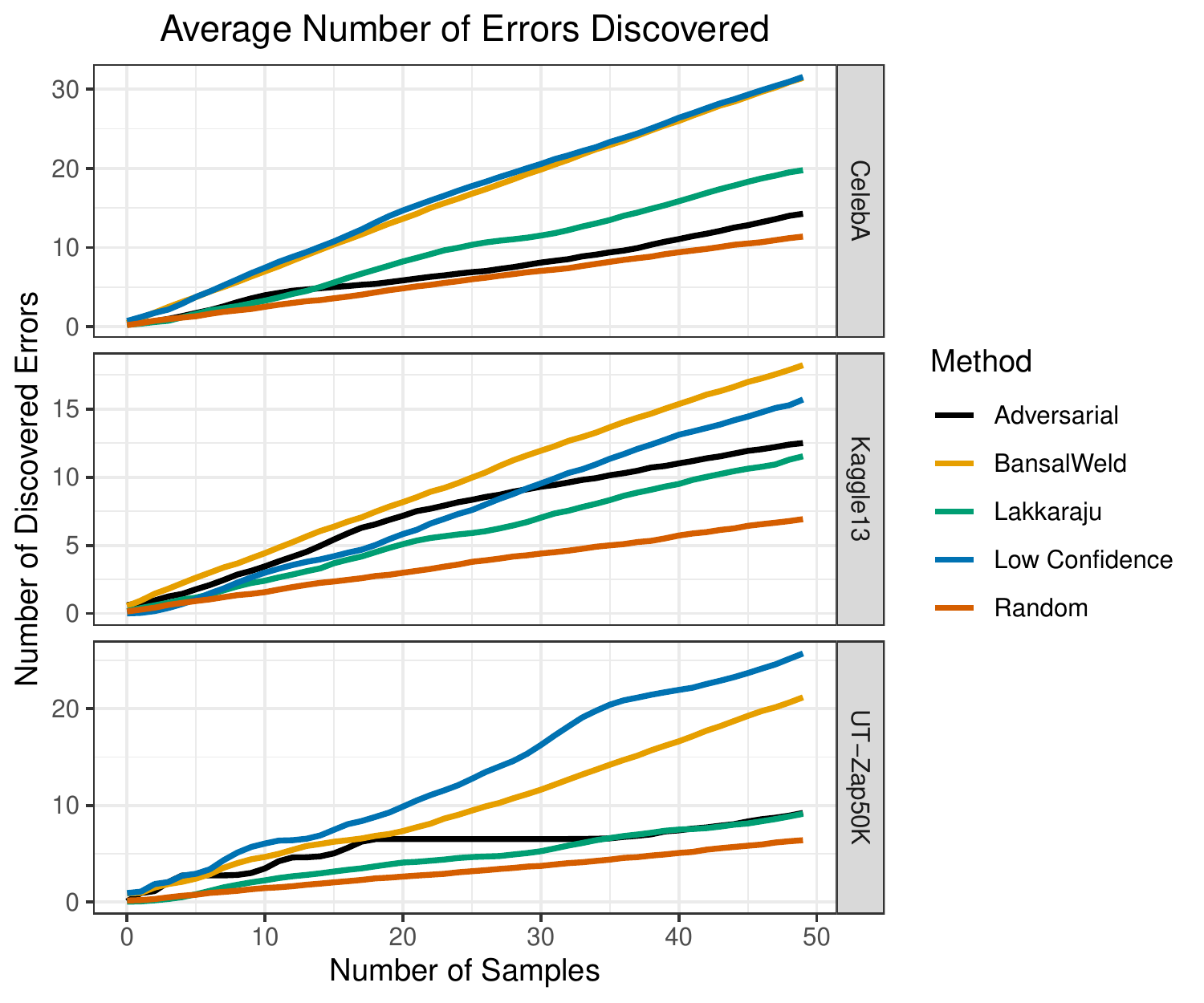}
	\end{center}
	\caption{Mean number of errors discovered  across 1,000 runs of the search methods.  This is the utility function presented by Lakkaraju for imagery.}
	\label{fig:fix}
\end{figure}

\begin{figure}[!htb]
	\begin{center}
		\includegraphics[width=0.9\linewidth]{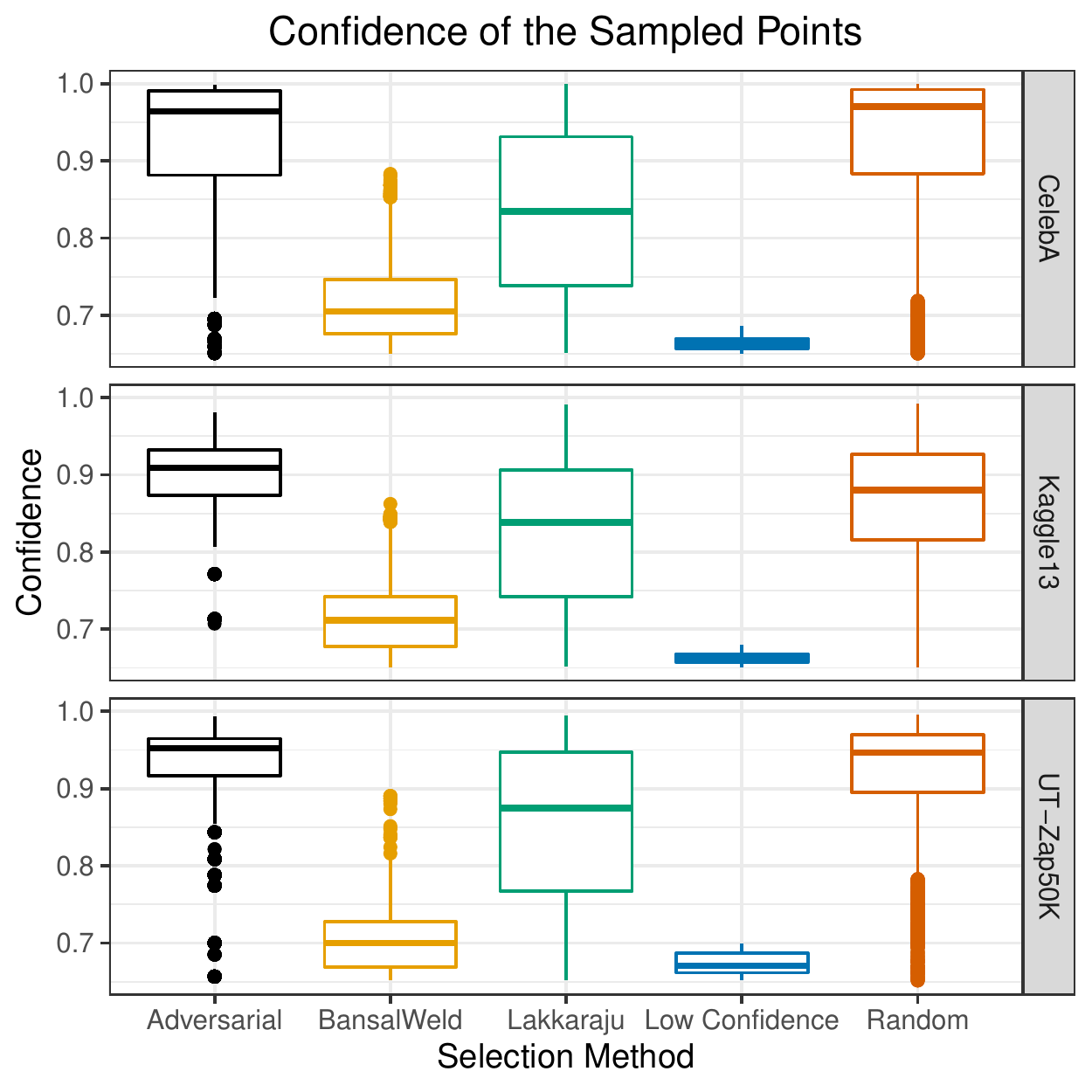}
	\end{center}
	\caption{Box plot showing the model confidence of the sampled data points.}
	\label{fig:con}
\end{figure}

The Bansal and Weld Utility function is defined in Equation \ref{eq:bw}, and shows that the utility function rewards the discovery of errors that occur near high-confidence points.  Being \emph{near} high-confidence points is an important distinction because it does not directly reward finding high-confidence errors.  Figure \ref{fig:bw} shows that the Bansal and Weld search achieves high values for the Bansal and Weld utility.  However, by looking at the number of errors discovered (Figure \ref{fig:fix}), and the confidence of the points sampled by the Bansal and Weld search (Figure \ref{fig:con}), it becomes obvious that high values of the Bansal and Weld utility can be achieved by finding a large number of lower confidence errors; even if these errors should be expected given the model confidence.  The Bansal and Weld search achieves an SDR similar to random search, and it is not clear that the search is achieving anything other than selecting samples in the lower confidence ranges.  This is further confirmed by the strong performance of the low confidence search for this utility function.  The Adversarial Distance search may not perform as well for this utility measure because it discovers fewer errors, but our results from the previous section show it still samples throughout the search space and finds more errors than expected given the confidences of the sampled predictions.  

The Lakkaraju utility function counts the number of errors discovered by the search method.  Figure \ref{fig:fix} shows that the low confidence search and the Bansal and Weld search maximize this utility.  However, as shown in Figure \ref{fig:con}, these methods sample lower confidence points, and so we should expect them to find errors at high  rates.  The Adversarial Distance search samples predictions with similar confidence levels to the random search (Figure \ref{fig:con}), but finds more errors.  We believe that this is strong evidence that our method finds errors that point to model overconfidence because both methods sample predictions of similar confidence, but the Adversarial Distance search finds more errors and achieves greater SDR values.

Interestingly, the Lakkaraju search method achieves competitive values for the Bansal and Weld utility while having a lower number of discovered errors.  It is likely that the discovered errors are in close proximity to high-confidence predictions.  Still, the query sets of the Lakkaraju search achieve a low SDR which indicates that errors are occurring at the expected rate (with the exception of the UT-Zap50K dataset which is revealed to have a large amount of overconfidence in Figure \ref{fig:overconfidence}).

\subsection{Discovered Errors}

\begin{figure}[!htb]
	\begin{center}
		\includegraphics[width=1\linewidth]{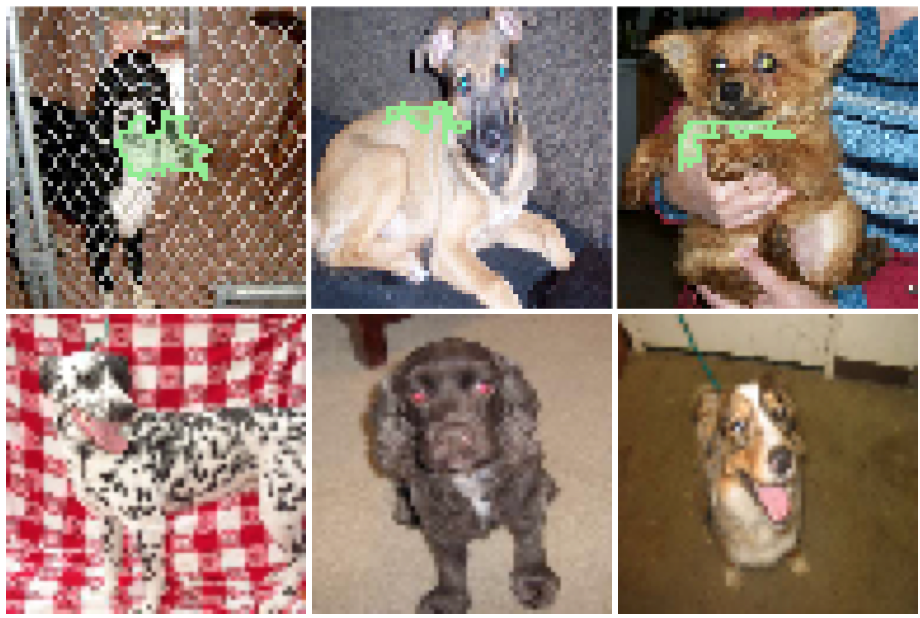}
	\end{center}
	\caption{Dogs predicted to be cats with high confidence.  Notice the dogs have light fur or are on a light background.  LIME also indicates that light colored superpixels are the most important indicator of the cat prediction (LIME did not identify critical superpixels for each image).}
	\label{fig:blindspotCat}
\end{figure}

Figure \ref{fig:blindspotCat} shows the first six errors discovered by the Adversarial Distance search for the Kaggle13 dataset.  LIME \cite{Ribeiro2016} has been run for each image to find the superpixels that the model considers most important in classifying these images as "cat".  Note that some images are missing LIME information (the green outline) because the method did not identify superpixels for that image that exceeded the default threshold of importance.  In general, the errors discovered by the Adversarial Distance search were of dogs with light fur, or dogs on a light background.  For the cases where LIME discovered important superpixels, the light-colored superpixels were the most important indicators of the image containing a cat.  These high-confidence mistakes suggest that the model is biased to place images with light colors into the cat class.  This is consistent with the training set containing only light furred cats after the dataset was biased.

Similar results can be found for the CelebA and UT-Zap50K datasets, but were not included for brevity.

\subsection{Insight to Adversarial Distance}

\begin{figure}[t]
	\begin{center}
		\includegraphics[width=1\linewidth]{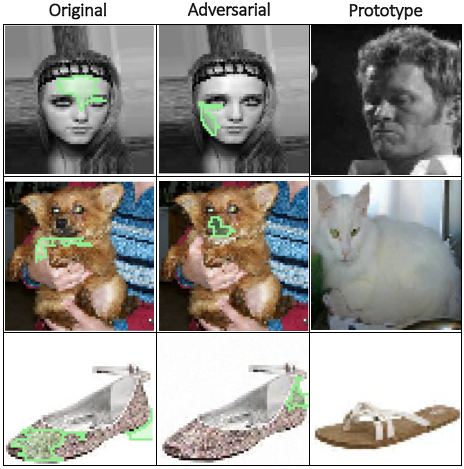}
	\end{center}
	\caption{The first column is the original image (incorrectly labeled the critical class) with LIME activation .  The second column is the adversarial image (now correctly classified) with LIME activation .  The third column is the image from the critical class with the highest Adversarial Distance.  It is interpreted as a prototypical instance from the critical class.}
	\label{fig:adversarial}
\end{figure}

Figure \ref{fig:adversarial} provides some insight as to how Adversarial Distance helps find high-confidence mistakes.  The first column shows the original image and the important superpixels leading to the image's misclassification.  The second column shows the adversarial image and the superpixels leading to the image's correct classification.  The third column shows the image from the critical class with the highest Adversarial Distance, as a kind of prototypical instance.

The first row of Figure \ref{fig:adversarial} is from the CelebA dataset.  For the original image, the classifier predicts the image to be "male", and may be focused on the absence of bangs.  After perturbing a very small number of pixels, the classifier predicts female and seems to be focused on the absence of sideburns (as shown in the prototypical male image).  In the second row, the classifier predicts the original image to be a cat, and seems focused on the light color of the hand (similar to the light color of the prototypical cat image).  In the adversarial image, the classifier predicts the image to be a dog, and is now focused on the dark nose.  In the third row, the classifier predicts the image to be "not shoe", and seems to be focused on the toe, and the absence of a heal.  For the adversarial image, the classifier predicts shoe and highlights the back of the shoe (the prototypical "not shoe" has no back).  

In the cases highlighted above, the classifier is incorrect in its prediction because it seems to focus on non-robust features of the image.  However, robust features that could lead to the correct prediction are also present in the image.  For example: the shoe has a well defined back, the dog has a dark nose, and the woman does not have sideburns or facial hair.  These images likely have low Adversarial Distances because these robust features exist in the image and only small perturbations are required to break the non-robust features leading to an incorrect prediction.  Additionally, because these robust features exist in the image, the classifier should not have been as confident in its prediction as it was.  A low Adversarial Distance seems to indicate the presence of contradictory robust features or model overconfidence.  Sampling these types of images helps discover errors at rates exceeding expectation.

\section{Conclusions}\label{sec:Conclusions}

In this work, we introduced the concept of Adversarial Distance and showed how it can be used to help discover prediction errors at rates exceeding what would be expected given the confidence of the model's predictions.  That is, when the Mean Absolute Error between an image and its adversarial version is lower than expected for a given classifier confidence, the classifier may be more confident in its prediction than is appropriate.  

Experimental results compared the Adversarial Distance search to existing methods designed to search for high-confidence classification errors.  Results showed that all methods achieved similar values of "spread", meaning, they all searched evenly throughout the problem's derived feature space.  However, the Adversarial Distance search achieved the largest Standardized Discovery Ratios, meaning, it resulted in the highest rate or error discovery relative to the expected error rate.

Future work should focus on the observation that the Adversarial Distance search seems to discover fewer errors as the number of search queries increases.  This is likely because the density of mistakes decreases as Adversarial Distance increases.  Therefore, when considering large searches, we believe it may be beneficial to use the Adversarial Distance search to prime methods that learn a meta-model of classifier error.  Additionally, future work should focus on the uses of the discovered errors.  For example, further model calibration or improved risk mitigation strategies.

\bibliographystyle{IEEEtran}
\bibliography{library}

\end{document}